\documentclass[10pt,twocolumn,letterpaper]{article}

\usepackage{cvpr}
\usepackage{times}
\usepackage{epsfig}
\usepackage{graphicx}
\usepackage{amsmath}
\usepackage{amssymb}

\usepackage{multirow}
\usepackage{bm}

\usepackage[breaklinks=true,bookmarks=false]{hyperref}

\cvprfinalcopy 

\def\cvprPaperID{18}

\begin{document}

\title{Pay Attention to Virality: understanding popularity\\ of social media videos with the attention mechanism
\vspace{-.2cm}}

\author{Adam Bielski\\
Tooploox\\
{\tt\small adam.bielski@tooploox.com}\vspace{-0.3cm}
\and
Tomasz Trzcinski\\
Warsaw University of Technology, Tooploox\\
{\tt\small tomasz.trzcinski@tooploox.com}\vspace{-0.3cm}
}

\maketitle

\begin{abstract}
Predicting popularity of social media videos before they are published is a challenging task, mainly due to the complexity of content distribution network as well as the number of factors that play part in this process. As solving this task provides tremendous help for media content creators, many successful methods were proposed to solve this problem with machine learning. In this work, we change the viewpoint and postulate that it is not only the predicted popularity that matters, but also, maybe even more importantly, understanding of how individual parts influence the final popularity score. To that end, we propose to combine the Grad-CAM visualization method with a soft attention mechanism. Our preliminary results show that this approach allows for more intuitive interpretation of the content impact on video popularity, while achieving competitive results in terms of prediction accuracy. 
\end{abstract}

\vspace{-0.4cm}
\section{Introduction}
Multiple factors make popularity prediction of social media content a challenging task, including propagation patterns, social graph of users and interestingness of content. 
Current methods for online content popularity analysis focus mostly on determining its future popularity~\cite{Khosla14,Chen16,Pinto13,Tatar14}. For instance, \cite{Khosla14} and~\cite{deza2015virality} use visual cues to predict the popularity of online images. \cite{Trzcinski17,TrzcinskiSVR,Stokowiec17} use machine learning methods, such as Support Vector Regression and recurrent neural networks, applied to visual and textual cues to predict the popularity of social media videos. 
\cite{Chen16} combines cues from different modalities for the same purpose. Although popularity prediction is an important problem, we believe that it does not address all the challenges faced by online video creators. More precisely, it does not allow to answer the question of many creators: how a given frame or title word contributes to the popularity of the video? Even though correlation does not mean causality, this kind of analysis allows to understand the importance of a given piece of content and prioritize the creation efforts accordingly. 

\begin{figure}[t]
\begin{center}
\includegraphics[width=0.95\linewidth]{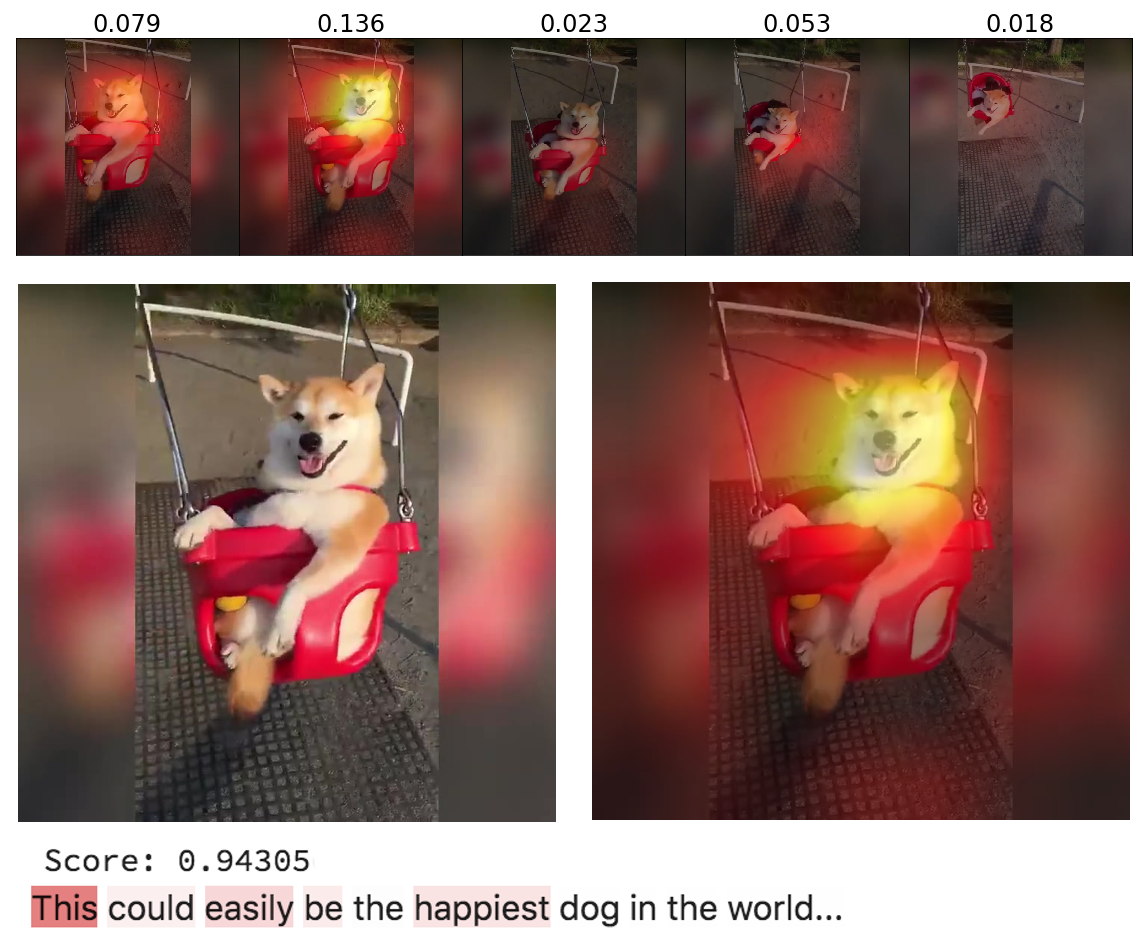}
\end{center}
\vspace{-12pt}
\caption{Sample social media video frames and its headline with visualized importance of their parts on predicted video popularity.  The {\bf top} row shows 5 consecutive frames of a social media video with their attention weights above. We use those weights to scale the magnitudes of Grad-CAM~\cite{gradcam} heatmaps when visualizing the importance of video elements on the popularity score. The {\bf bottom} row shows the frame with the highest attention weight {\bf (left)} with its popularity importance visualization {\bf (right)}. For the headline text darker color corresponds to higher importance.}
\vspace{-0.5cm}

\label{fig:vis}
\end{figure}

In this paper, we outline a fundamentally different approach to online video popularity analysis that allows social media creators both to predict video popularity as well as to understand the impact of its headline or video frames on the future popularity. To that end, we propose to use an attention-based model and gradient-weighted class activation maps \cite{gradcam}, inspired by the recent successes of the attention mechanism in other domains~\cite{ShowAttendTell,Yang2016HierarchicalAN}. Although some works focused on understanding the influence of image parts on its popularity~\cite{Khosla14,viraliency}, our method addresses videos, not images, and exploits the temporal characteristics of video clips through the attention mechanism. 
By extending the baseline popularity prediction method with the attention mechanism, we enable more intuitive visualization of the impact visual and textual features of the video have on its final popularity, while achieving state-of-the-art results on the popularity prediction task.

\section{Popularity prediction with attention}

We cast the problem of social media video popularity prediction as a binary classification task, as in~\cite{Trzcinski17,Stokowiec17}. We focus on videos from Facebook and normalize their viewcount by the number of page followers. We assign a popular/unpopular label by splitting our dataset at the median normalized viewcount, following the approach of~\cite{SimoSerra2015}. We use a cross-entropy loss function to classify a video as popular/unpopular and take a set of video frames and/or headline features as an input.

\textbf{Video frames.} We extract $N=18$ evenly distributed frames from the first 6 seconds of a video\footnote{We use the first seconds of a video as this is how Facebook counts views, but we can extend our method to longer videos through sampling.}. We use 2048-dimensional output of the penultimate layer of ResNet50~\cite{He15} pre-trained on ImageNet~\cite{imagenet}  to get a high-level frame representation as in~\cite{Trzcinski17}. For each frame feature vector we apply a learnable linear transformation followed by ReLU, obtaining a sequence of frame embeddings $(\bm{q}_j)^{N}_{j=1}$. The final video embedding is a weighted average of these embeddings $\bm{v} = \sum_{i=1}^{N} \alpha_i \bm{q}_i$. 
Weights $\alpha_i$ are computed with attention mechanism implemented as a two-layer neural network~\cite{Yang2016HierarchicalAN}: the first layer produces a hidden representation $\bm{u}_i = \tanh(\bm{W}_u \bm{q}_i + \bm{b}_u)$ and the second layer outputs unnormalized importance $a_i = \bm{W}_a \bm{u}_i + b_a$. $\bm{W}_a$ can be interpreted as a trainable high level representation of {\it the most informative vector} in $\bm{u}_i$ space. 
Final weights are normalized with softmax: $\alpha_i=\exp(a_i)/\sum_k \exp(a_k)$.

\textbf{Headline.} We represent a headline as a sequence of pre-trained GloVe \cite{glove} word vectors $(w_t)^N_{t=1}$. We handle sequences of variable length using a bidirectional LSTM. Similarly to video frames, we use a two-layer attention mechanism on hidden state vectors $\bm{h}_t$ to let the network learn the importance coefficients $\beta_t$ for each word. The final text representation $\bm{d}$ is a weighted average of hidden state vectors $\bm{d}=\sum_{t=1}^N\beta_t\bm{h}_t$.

\textbf{Multimodal prediction.} We concatenate previously trained video and text embeddings and train a two-layer neural network for popularity prediction. The intermediate layer output serves as multimodal embedding that captures information from both image and text modality that contribute to image popularity.

\textbf{Visualizations.} We visualize the importance of visual features using Grad-CAM \cite{gradcam}. More precisely, we generate heatmaps pointing to regions contributing to popularity in each frame. To this end, we compute gradients of the popular class score $\hat{s}$ with respect to the output of the last convolutional layer of ResNet50 $A \in R^{K \times K \times F}$. Gradients are then used to compute weights $\gamma_f = \frac{1}{K^2} \sum_{i, j = 1}^K \frac{\partial \hat{s}}{\partial A_{i,j}^f}$ that applied to the convolutional output create class activation map $\bm{H} = \max(0, \sum_{f=1}^F \gamma_f A^f)$. We then normalize the heatmap values to [0, 1] and use attention weights to scale the heatmap by $\alpha_i/\max(\alpha)$. This way we obtain a sequence-wide normalized heatmap of frame regions influencing the final popularity score.

For visualizations in the text domain, we use attention weights $\beta_t$ used to compute text representation $\bm{d}$. These weights capture relative importance of words in their context to headline popularity, as shown in~\cite{Yang2016HierarchicalAN} in the context of sentiment analysis.

\section{Experiments}

\begin{table}[t!]
\begin{center}
\setlength\tabcolsep{1.6pt}
\def\arraystretch{0.9}
\begin{tabular}{|c|c|c|c|}
\hline
Input & Features & Acc [\%] & Spearman \\
\hline\hline
\multirow{2}{*}{Video frames} & ResNet50 mean & 68.17 & 0.524 \\ \cline{2-4}
& + attention& 68.87 & 0.526 \\
\hline
\multirow{2}{*}{Headline} & biLSTM~\cite{Stokowiec17} & 69.47 & 0.542 \\ \cline{2-4}
 & + attention & 68.70 & 0.525 \\
\hline
\multirow{2}{*}{Multimodal} & ResNet + biLSTM & 71.94 & {\bf 0.612} \\ \cline{2-4}
& + attention & {\bf 72.72} & 0.607 \\
\hline
\end{tabular}
\end{center}
\vspace{-5pt}
\caption{Video popularity prediction results.
}
\vspace{-0.6cm}
\label{tab:results}
\end{table}

We use a dataset of 37k Facebook videos with 80/10/10 train/validation/test splits. We use validation set to 
perform randomized serach of hyperparameters such as embedding dimensionalities, dropout rates and batch normalization use. 
For training headline embeddings we use frozen pre-trained GloVe word vectors trained on Wikipedia and Gigaword~\cite{glove}. As baselines, we use a simple mean of ResNet50 feature vectors as input to two layer neural network (video frames) and concatenation of last states of LSTM  (headlines).
We use Keras for implementation. 
\textbf{Results.}
For all methods, we follow the evaluation protocol of~\cite{Trzcinski17,Stokowiec17} and compute the classification accuracy and Spearman correlation between the predicted probability of popular label and normalized view count of a video. 
Tab.~\ref{tab:results} shows the results. Interestingly, almost equal popularity prediction results can be obtained using either video frames or headline features. Combining both modalities leads to noticeable improvement, while adding attention mechanism improves the performance in the multimodal and visual case. For headlines, the performance with attention deteriorates slightly. We speculate that the bi-directional LSTM already learns internal dependencies between hidden states and adding attention cannot help further, while for video frames it enables the network to exploit the temporal dependencies between the frames. 

Overall, we see that prediction methods with attention achieve competitive results, while thanks to the attention mechanism we can increase the interpretability of our model. 
As Fig.~\ref{fig:vis} shows, extending standard Grad-CAM visualization with the attention mechanism allows us  to interpret the influence of each video frame.  Although the preliminary results presented in this paper leave place for improvement, they indicate the potential of using attention mechanism to increase the interpretability of popularity prediction methods for social media videos.

\clearpage

{\small
\bibliographystyle{ieee}
\bibliography{egbib}
}

\end{document}